\documentclass[letterpaper, 10 pt, conference]{ieeeconf}

\IEEEoverridecommandlockouts 

\usepackage{graphicx}
\usepackage{amsmath,amssymb,amsfonts}
\usepackage{mathtools}
\usepackage{algorithm}
\usepackage{algpseudocode}
\usepackage{subcaption}
\usepackage{multirow}
\usepackage{booktabs}
\usepackage{dsfont}
\usepackage{array}
\usepackage{wrapfig}
\usepackage{xcolor}

\let\labelindent\relax %
\usepackage{enumitem}
\usepackage{lipsum}
\usepackage{hyperref}
\usepackage{cleveref}
\usepackage{bbm}
\usepackage{times}
\usepackage{caption}
\usepackage{pifont}
\newcommand{\cmark}{\ding{51}}
\newcommand{\xmark}{\ding{55}}

\usepackage[backend=biber,
            style=ieee,
            natbib=true, 
            url=false,
            eprint=false,
            doi=false]{biblatex} 
\definecolor{hiroshige}{HTML}{ffd06f}
\definecolor{darkBlue}{RGB}{10,50,220}
\definecolor{customRed}{RGB}{190,110,113}
\definecolor{customGreen}{RGB}{70,170,80}

\hypersetup{
    breaklinks=true,letterpaper=true,colorlinks,citecolor=blue
}
\title{\raisebox{-0.6ex}{\includegraphics[height=8mm]{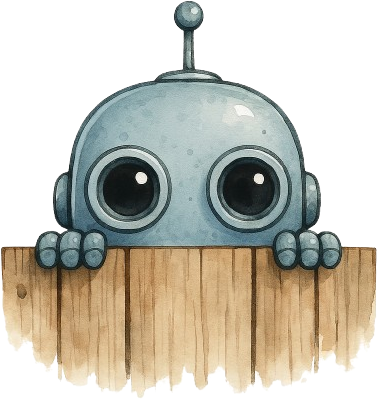}} \LARGE \bf
\method: Guiding and Minimal Image Representations\\ for Zero-Shot Generalization of Robot Manipulation Policies
}

\author{
Jesse Zhang$^{\star 1,2,3}$, Marius Memmel$^{\star 1,2}$, Kevin Kim$^{3}$, \\
Dieter Fox$^{1,4}$, Jesse Thomason$^{3}$, Fabio Ramos$^{2}$, Erdem Bıyık$^{3}$, Abhishek Gupta$^{\dag 1}$, Anqi Li$^{\dag 2}$%
\thanks{$^\star$Co-first authors, $^\dag$Equal Advising, $^{1}$University of Washington, $^{2}$NVIDIA, $^{3}$University of Southern California, $^{4}$Allen Institute for AI}
}

\begin{document}
\newcommand{\method}{PEEK}
\newcommand{\methodlong}{PEEK (\textbf{P}olicy-agnostic \textbf{E}xtraction of \textbf{E}ssential \textbf{K}eypoints)}
\newcommand{\todo}[1]{\textcolor{red}{TODO: #1}}
\newcommand{\datapolicy}{\mathcal{D}_\pi}
\newcommand{\datavlm}{\mathcal{D}_\text{VLM}}
\newcommand{\peekobs}{o^{p,m}}
\newcommand{\peekobsvlm}{o^{p,m}}
\newcommand{\task}[1]{\textsc{#1}}
\newcommand{\baseline}[1]{\texttt{#1}}
\maketitle

\thispagestyle{empty}
\pagestyle{empty}

\begin{abstract}
Robotic manipulation policies often fail to generalize because they must simultaneously learn \emph{where} to attend, \emph{what} actions to take, and \emph{how} to execute them. We argue that high-level reasoning about \emph{where} and \emph{what} can be offloaded to vision-language models (VLMs), leaving policies to specialize in \emph{how} to act. We present \textbf{\methodlong}, which fine-tunes VLMs to predict a unified point-based intermediate representation: (1) end-effector paths specifying \emph{what} actions to take, and (2) task-relevant masks indicating \emph{where} to focus. These annotations are directly overlaid onto robot observations, making the representation policy-agnostic and transferable across architectures. To enable scalable training, we introduce an automatic annotation pipeline, generating labeled data across 20+ robot datasets spanning 9 embodiments. In real-world evaluations, \method\ consistently boosts zero-shot generalization, including a \textbf{41.4$\times$} real-world improvement for a 3D policy trained only in simulation, and \textbf{2--3.5$\times$} gains for both large VLAs and small manipulation policies. By letting VLMs absorb semantic and visual complexity, \method\ equips manipulation policies with the minimal cues they need---\emph{where}, \emph{what}, and \emph{how}. 
Website at \url{https://peek-robot.github.io}.
\end{abstract}

\section{Introduction}

Imagine walking through a crowded store when your child suddenly cries out, ``I want the Labubu!''
Though you’ve never heard the word before, context clues guide your eyes to the fuzzy toy on the shelf, and you effortlessly weave through the crowd to grab it. 
What makes this possible is not raw perception ability, but the ability to interpret ambiguous instructions and distill them into just the right cues---\emph{where} to focus, \emph{what} actions to take, and \emph{how} to perform these actions at the low level. Similarly, if given \emph{where} to focus and \emph{what} motions to take, a robot manipulation policy should be able to achieve the visual robustness and semantic generalization necessary for open-world deployment by focusing only on \emph{how} to perform actions. 

A common tactic for training manipulation policies is through imitation learning of human-collected robotics data~\citep{zhao23aloha, kim2024openvla, black2024pi0, Yan2025ManiFlow}, which attempts to learn the where, what, and how all at the same time.
Yet their performance degrades on novel objects, clutter, or semantic variations~\citep{taxonomy2025arxiv, atreya2025roboarena}, since the policy alone bears the burden of handling task, semantic, and visual complexity. 
Such failures often entangle the axes of \emph{where}, \emph{what}, and \emph{how}---for example, grasping a distractor simultaneously reflects misplaced attention, an incorrect object choice, and a wrong motion.

\begin{figure}[t]
    \centering
    \includegraphics[width=\linewidth]{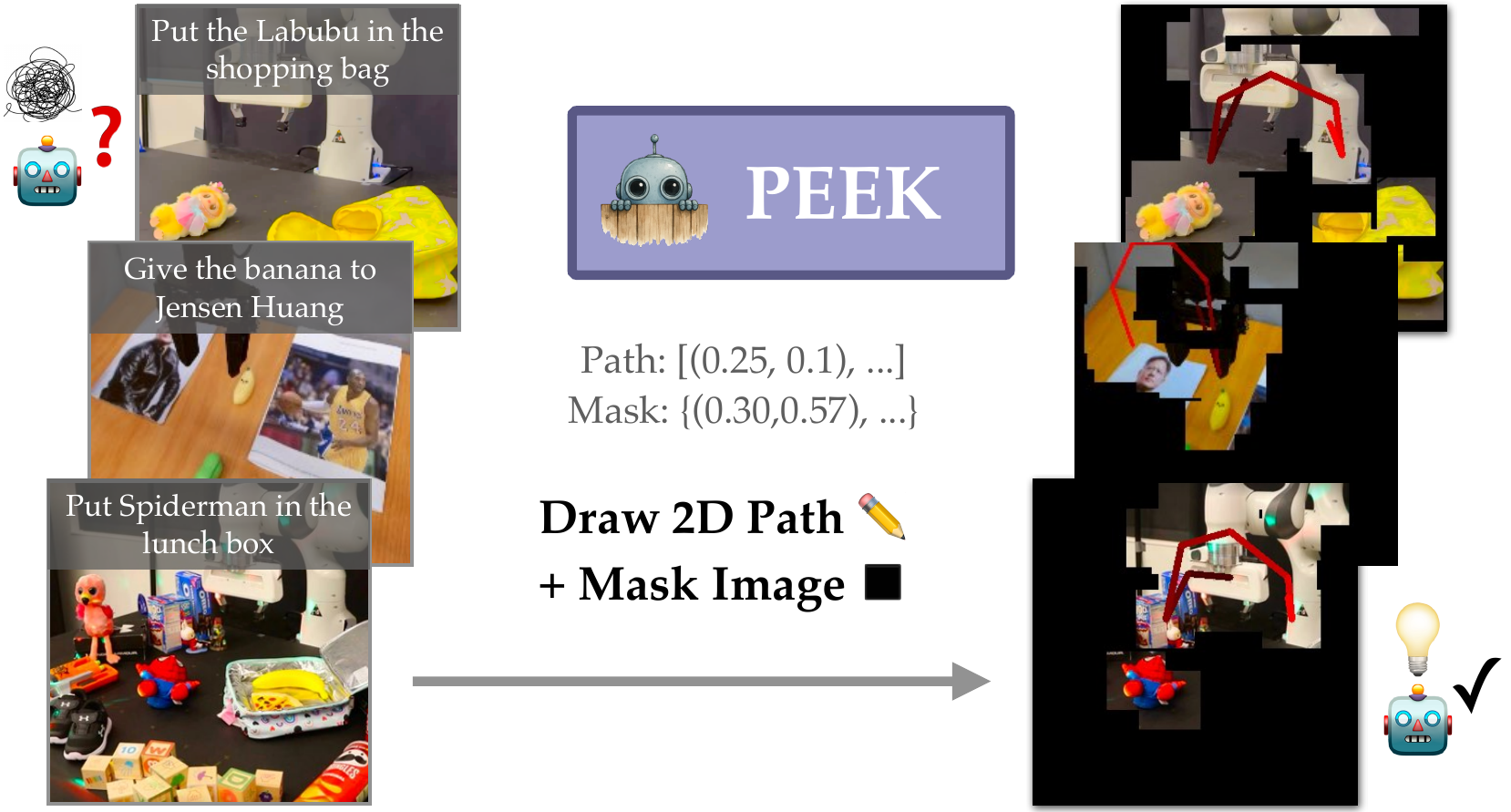}
    \caption{\footnotesize{\method\ enables policy generalization by modulating minimal representations of \emph{where} to focus and \emph{what} to do for robust policy learning.}}
    \label{fig:teaser}
    \vspace{-2em}
\end{figure}

Our key idea is to offload high-level reasoning to vision-language models (VLMs), which can excel at semantic and visual generalization~\citep{li2025hamster, molmoact2025}, leaving the policy to determine how low-level behavior should be executed. Instead of forcing the policy to directly parse raw images and instructions, a high-level VLM modulates the input representation to the low-level policy by providing: (1) a path that encodes \emph{what} the policy should do, and (2) masks showing \emph{where} to attend. By ``absorbing'' semantic and visual variation, the VLM provides the policy a simplified, annotated ``peek'' of the scene that gives the what and the where, while the policy only needs to learn \emph{how} to perform the low-level actions. This intermediate representation helps policy execution inherit many of the VLM's semantic and visual generalization capabilities. Our VLM-modulated representation is naturally policy-agnostic, allowing it to be applied to arbitrary image-input robot manipulation policies, including state-of-the-art RGB and 3D manipulation policies~\cite{ke20243d, zhao23aloha, black2024pi0}.

To concretely instantiate this insight into a practical algorithm, we introduce \methodlong, which proposes a unified, point-based intermediate representation that trains VLMs to predict \emph{what} policies should do and \emph{where} to focus on. %
Specifically, we propose to finetune pretrained VLMs~\cite{vila2024} to predict a sequence of points corresponding to (1) a \emph{path} that guides the robot end-effector in what actions to take and (2) a set of task-relevant \emph{masking points} that show the policy where to focus on (see \Cref{fig:teaser}). During low-level visuomotor policy training and inference, we modulate the policy's image observations by directly drawing these VLM-predicted paths and masks onto the image, allowing the policy to simply focus on how to act, rather than learning all three simultaneously. Doing so significantly bolsters policy generalization, combining the generality of high-level VLM predictions with the precision of low-level policy learning. In this paper, we instantiate a full-stack implementation of \method, from devising a scalable data annotation scheme that enables large-scale VLM finetuning on robotic datasets to representation-modulated training of low-level robot policies from simulation and real world data.

In 535 real-world evaluations across 17 task variations, we demonstrate that \method\ consistently boosts zero-shot policy generalization: a 3D policy (3DDA~\citep{ke20243d}) trained only in simulation achieves \textbf{41.4$\times$} higher success in the real world when guided by \method, and both large-scale vision-language-action models ($\pi_0$~\citep{black2024pi0}) and small transformer-based policies~\citep{zhao23aloha} see \textbf{2–3.5$\times$} success rate improvements. These results demonstrate the power of using high-level VLMs to absorb task complexity, providing low-level policies with exactly the minimal cues they need for generalizable manipulation.

\section{Related Works}
\label{sec:related}

\noindent\textbf{Object-Centric Representations.}
One approach to improving the visual generalization of imitation learning (IL) policies is to build object-centric representations~\citep{shi2024plug, emukpere2025docir, mirjalili2025augmentedrealityrobotsarro, hancock2025byovla, yuan2025roboengine, huang2025otter, li2025controlvla}.
Earlier works relied on human-selected abstractions or manual annotation~\citep{shi2024plug}, while more recent methods leverage pre-trained, open-vocabulary segmentation models to visually isolate task-relevant objects~\citep{emukpere2025docir, mirjalili2025augmentedrealityrobotsarro, hancock2025byovla, yuan2025roboengine, li2025controlvla}.
Among these, ARRO~\citep{mirjalili2025augmentedrealityrobotsarro} is closest to our work, proposing a policy-agnostic masking scheme using GroundingDINO~\citep{liu2023grounding} to filter images for task-relevant objects.
However, we found in \Cref{sec:experiments} that such object detectors often fail in cluttered, realistic scenes. %
By contrast, \method\ queries a fine-tuned VLM to predict task-relevant masking points directly, resulting in more robust masks than using object-detection models due to the VLM's extensive pre-training.
Another approach, OTTER~\citep{huang2025otter}, implements \emph{implicit} masking by filtering CLIP image patches, but this approach is architecture specifc. 
\method's policy-agnostic explicit masking allows us to integrate it with far more powerful policy backbones than OTTER, i.e., vision-language-action models like $\pi_0$~\citep{black2024pi0}.
Finally, while masking alone helps mitigate visual distractors, it alone cannot handle semantic variation; \method\ also provides explicit action guidance via predicted paths.

Another line of object-centric methods relies on \emph{learning} to decompose scenes into object-level representations in a self-supervised manner, e.g., via slot-attention~\citep{locatello2020slotattention, biza23invariant, zhang2023slotoptimaltransport, sold2025mosbach}, which learns to map visual features into a set of discrete, object-centric ``slots'' through competitive attention mechanisms.
However, these methods have not been applied to real-world robot manipulation settings and generally do not work zero-shot.
\method's use of a pre-trained VLM helps it predict task-relevant points on new objects and tasks.

\noindent\textbf{Guiding Manipulation Policies.}
A separate line of work improves generalization by explicitly guiding policies in \emph{how} to perform tasks via 2D gripper paths. 
RT-Trajectory introduced this concept using human-drawn sketches at inference time~\citep{gu2023rttrajectory}. 
Later methods integrated 2D path prediction into VLA training objectives~\citep{niu2024llarva, huang2025thinkact, molmoact2025, zheng2025tracevla}, but these approaches are tied to specific architectures.
More relevant is HAMSTER~\citep{li2025hamster}, which trains a VLM to predict future 2D gripper paths that a lower-level 3D policy conditions on.
While this approach aids with policy understanding of \emph{what} high-level motions to perform, we found in \Cref{sec:experiments} that HAMSTER-trained policies are easily confused by visual variation.
In contrast, \method's VLM predicts a single point-based representation that also includes masks helping the policy understand \emph{where} to focus on.

Other works propose guiding policies via relabeling language instructions~\citep{xiao2022robotic, zhang2023bootstrap, zhang2023sprint, smith2025steer, Chen25-ecot-lite} or behavioral priors, i.e., latent \emph{skills}, learned from data~\citep{pertsch2020spirl, singh2021parrot, ajay2021opal, zhang2024extract}.
These approaches are complementary to \method's image-based input representation.

\section{\method: Guiding and Minimal Image Representations}
\label{sec:method}
We study how to enhance the generalization capability of arbitrary visuomotor policies to semantic and visual task variation.
To do so, \method\ proposes to offload high-level task reasoning to VLMs to produce a \emph{guiding} (what) and \emph{minimal} (where) image representation for a low-level policy, which in turn actualizes \emph{how} to actually perform the task through real-world actions. Concretely, we instantiate this representation via 1) 2D gripper paths and 2) task-relevant masks (see \Cref{fig:teaser}). This hierarchical approach shifts the burden of generalization from the low-level policy to the high-level VLM,
allowing the policy to focus only on \emph{how} to execute low-level actions.

\subsection{Conceptual Insight}

Imitation learning methods train a policy $\pi(a_{t} \mid o_t, s_t, l)$ predicting an action $a_{t}$ given RGB observations $o_t$, proprioceptive and other sensory data (e.g., depth) $s_t$, and a task instruction $l$.
Given an expert-collected robot dataset $\datapolicy$, %
$\pi$ is trained with maximium likelihood estimation, i.e., $\max_{\pi}\mathbb{E}_{(o_t, s_t, l, a_{t})\sim \datapolicy} \left[\log \pi(a_{t} \mid o_t, s_t, l) \right]$.

\method\ explores how to improve imitation learning methods by training a VLM to map $(l, o_t)$ to a \textit{guiding} but \textit{minimal} representation, $\peekobs_t$, that enables zero-shot generalization to significant visual and semantic variation beyond that in $\datapolicy$. 
Downstream task variation can include any combination of, e.g., new scenes, visual clutter not present during training, new objects, and unseen language instructions.

Formally, \method\ fine-tunes a pre-trained VLM conditioned on $(l, o_t)$ to produce a set of points, i.e., $p_t, m_t \sim \text{VLM}(\cdot \mid o_t, l)$, corresponding to: (1) 2D gripper paths, $p_t$, indicating where the end-effector should move to solve the task, and (2) a set of task-relevant masking points, $m_t$, that indicate objects and regions of relevance.
2D gripper paths are defined as $p_t = [(x, y)_t, ..., (x,y)_T]$ where $(x, y) \in [0, 1]^2$ are normalized pixel locations of the end effector's positions at timestep $t$ until trajectory end point $T$.
Masking points are defined as $m_t = \{(x, y)_i\}_{i=1}^M$, an unordered set of pixel locations $(x, y) \in [0, 1]^2$ of task-relevant points.

Although any pre-trained text and image input VLM can be used to predict these path and mask points, prior work has found that even the best closed-source models struggle with predicting robot gripper paths without fine-tuning~\citep{molmoact2025, li2025hamster}, let alone masking points.
Therefore, we need to \emph{fine-tune} a VLM on a large dataset that grounds it to a diverse set of robot scenes and embodiments.
\method\ introduces a scalable data-labeling scheme which we use to create a dataset of over 2M VQA pairs, spanning 148k trajectories, 9 embodiments, and 21 robotics datasets.

\begin{figure*}[ht]
    \centering
    \includegraphics[width=\linewidth]{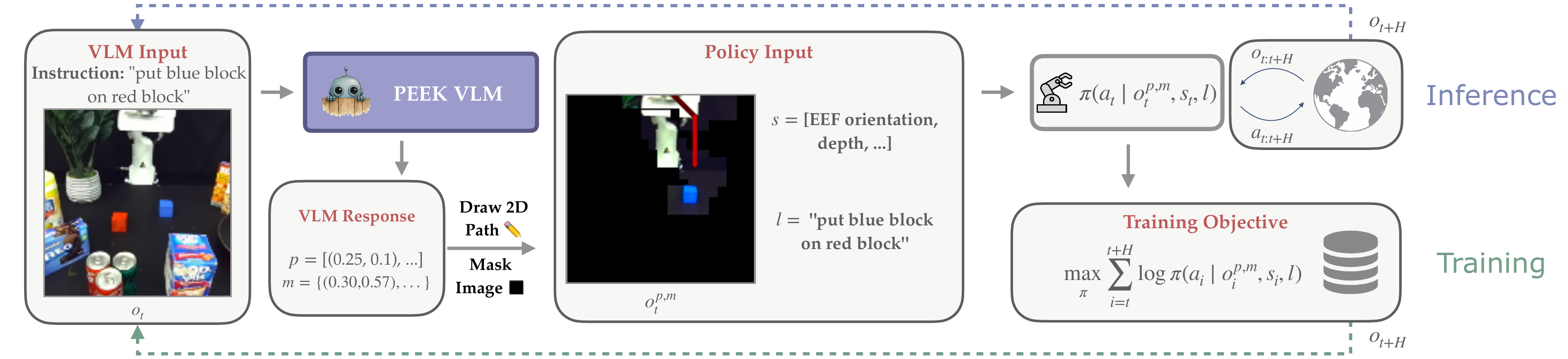}
    \caption{\footnotesize{\textbf{Policy Training and Inference Pipeline.} The VLM is called every $H$ steps to generate a path and task-relevant points. An (arbitrary) RGB-input policy is conditioned on the path and masked image to either predict actions for \textcolor[HTML]{7D87B2}{inference} or for \textcolor[HTML]{749E89}{training}. The same path and mask is applied onto incoming observations for $H$ steps, after which the VLM is re-queried.}}
    \label{fig:policy training and inference}
\end{figure*}

\subsection{VLM Data Preparation}
\label{sec:method:data labeling}
To finetune VLMs for \method\, we assemble a dataset $\datavlm = \{ (o, l, \texttt{ans})_i\}_{i=1}^V$ of image inputs $o$, instructions $l$, and text-based responses $\texttt{ans}$ depending on the dataset. In this section, we introduce our datasets, and then we detail our automatic robot data labeling pipeline.

\noindent\textbf{Point Prediction and VQA Datasets.} Like prior work~\citep{li2025hamster, molmoact2025}, we first incorporate readily available pixel point prediction and visual question answering (VQA) data into $\datavlm$ to maintain the VLM's general world knowledge and object reasoning capabilities. 
We use the RoboPoint dataset~\citep{yuan2024robopoint} with 770k pixel point prediction tasks, e.g., $l = \text{``Point to the cushions,''}$ and $\texttt{ans} = [(0.56, 0.69), (0.43, 0.67)]$, and 665k VQA examples, e.g., $l = \text{``What is the cat eating?,''}$ and $\texttt{ans} = \text{``An apple.''}$

\noindent\textbf{Robotics Datasets.}
Our main robotics dataset comes from the Open X-Embodiment (OXE) dataset~\cite{open_x_embodiment_rt_x_2023}, where we label 20 datasets from the OXE ``magic soup''~\citep{kim2024openvla}.
Notably, our data labeling pipeline works effectively on datasets with lots of clutter or awkward viewpoints that make task-relevant objects appear very small, such as DROID~\citep{khazatsky2024droid} (e.g., the pen in \Cref{fig:data labeling}).
In contrast, we found the pre-trained object detection models~\citep{liu2023grounding} used by prior works to extract object-centric representations~\citep{mirjalili2025augmentedrealityrobotsarro, li2025controlvla} to be ineffective.
Finally, we also include a robotics simulation dataset (LIBERO-90~\citep{liu2023libero}) in our training mix to broaden the visual feature coverage of the VLM. Now we describe how we scalably label our robotics datasets.

\noindent\textbf{Automatically Labeling Robotic Datasets.} \method's VLM needs to predict a list of 2D gripper path points $p_t$ and task-relevant masking points $m_t$ given arbitrary task instructions and robot observations.
Prior works label their dataset using calibrated 3D cameras (in simulation and the real world) or human annotations~\citep{li2025hamster, molmoact2025}, limiting the scalability of data annotation.
In contrast, we devise an automatic and scalable multi-step tracking pipeline that extracts how to solve the task and what to focus on directly from robot videos.

First, our representation should be \textit{minimal}, i.e., it should encode task-relevant entities at each timestep $t$. To extract this information from a video, we have to ask the following question: \textit{What entities are relevant to the task?}
We answer this question by tracking a grid of points through time with a visual point tracking model~\citep{karaev2025cotracker}. 
Points that move significantly throughout the trajectory correspond to the robot arm or objects being manipulated. 
We define this set as \emph{task-relevant} points $P^{task}_{t}=\{(x,y)_i\}_{i=1}^N$, tracked across all timesteps of a trajectory $t\in[1, T]$, as they capture the minimal information needed by a policy to solve the task.

Second, our representation should be \textit{guiding}, i.e., capture information about the (1) future relevant object movement and (2) robot gripper movement. 
(1) The tracking points tell us the entities' position at each timestep $t$. To capture how they move and where they end up, e.g., object placement locations, we include points at the last timestep $P^{task}_T$.
(2) We additionally construct a set of \emph{end-effector} points $P^{grip}_t=[(x,y)]_t^T$ by tracking the gripper throughout the video.

Finally, we process the data into subtrajectories separated by when the robot manipulates an object, and construct the 2D paths $p_t=P^{grip}_t$ and masking points $m_t=P^{task}_t\cup P^{task}_T$.
The natural language prediction target \texttt{ans} for the VLM is then a combination of the shortened $p_t$ and $m_t$: \texttt{TRAJECTORY: [(0.25, 0.1), ...] MASK: [(0.30,0.57), ...]}.
See \Cref{apdx: data labeling} and \Cref{fig:data labeling} for details regarding the data labeling pipeline.

\subsection{VLM and Policy Training/Inference with \method\ }
\label{sec:method:policy}

\noindent\textbf{VLM Fine-tuning.}
We use \texttt{VILA-1.5-3b}~\citep{vila2024} as our base VLM, a 3B parameter VLM trained on interleaved image-text datasets and video captioning data. We fine-tune our VLM for one epoch using the combined datasets totalling 3.5M samples with a learning rate of $5e^{-2}$ and a batch size of 16. Fine-tuning takes $\sim20h$ on 8 NVIDIA A100 GPUs.
We fine-tune the VLM with a standard supervised prediction objective to maximize the log-likelihood of the answers \texttt{ans}: $\max_{\text{VLM}} \mathbb{E}_{(o, l, \texttt{ans}) \sim \datavlm} \log \text{VLM}(\texttt{ans} \mid o, l)$.

\noindent\textbf{VLM Inference.}
During deployment, \method's VLM acts at a higher level, absorbing the semantic complexity and visual clutter of the scene and providing a lower-level policy with a guiding and minimal representation. However, querying the high-level VLM at every timestep is unnecessary because the scene is unlikely to change significantly at the same frequency as the policy is acting. Since frequent VLM queries are expensive and must be run sequentially, we run the VLM at a reduced frequency.
While prior works predict paths either at the start of a rollout~\citep{li2025hamster} or at every timestep~\citep{niu2024llarva}, our hybrid approach strikes a balance between inference speed and responsiveness.
To minimize the gap between training and deployment, our data labeling and training scheme reflects this design choice by querying the VLM at a fixed frequency of every $H$ timesteps.

\noindent\textbf{VLM / Policy Interface.}
During inference, the policy receives an augmented image input $\peekobsvlm_{t}$ created by drawing the path $p_t$ and mask $m_t$ onto the image observation $o_t$.

We \emph{draw} the 2D path $p_t$ by connecting each subsequent point in $p_t$ with a colored line segment. This drawing guides the policy for which path to follow to accomplish the task.
To indicate passage of time, the line segment changes from dark to light red {\color[HTML]{400000}{\rule{0.7em}{0.7em}}}{\color[HTML]{700000}{\rule{0.7em}{0.7em}}}{\color[HTML]{a00000}{\rule{0.7em}{0.7em}}}{\color[HTML]{d00000}{\rule{0.7em}{0.7em}}}{\color[HTML]{FF0000}{\rule{0.7em}{0.7em}}}.
To create masks, we start from a black canvas and use the area around the predicted task-relevant points to reveal parts of the image.
For each predicted task-relevant point $(x, y) \in m$, we create a square centered around $(x, y)$ with edge length 8\% of the image's size.
See \Cref{fig:policy training and inference} for a visual depiction of path and mask drawing.

We query the VLM every $H$ steps to generate $p_t, m_t$ based on the current environment observation $o_t$ and apply the same annotations $p_t$ and $m_t$ to all incoming observations $\peekobsvlm_{t:t+H}$ until $H$ steps have passed.

Each VLM query takes about 4-6 seconds on an RTX 3090 without any explicit speed optimization, but until the next VLM query, the policy $\pi$ runs at its own inference speed.

\noindent\textbf{Policy Training.}
Consequently, we annotate all the trajectories in the policy training data $\datapolicy$ to create an annotated dataset $\datapolicy^{p, m}$.
We train $\pi$ on the \method-labeled dataset $\datapolicy^{p, m}$ using its original training objective, e.g., maximizing log-likelihood of the actions: 
$\max_{\pi} \mathbb{E}_{\datapolicy^{p, m}} \log \pi(a_t \mid \peekobsvlm_t, s_t, l).$
We list policy and VLM query frequencies in \Cref{sec:apdx:implementation details}.

\begin{figure*}[tb]
    \centering
    \includegraphics[width=\linewidth]{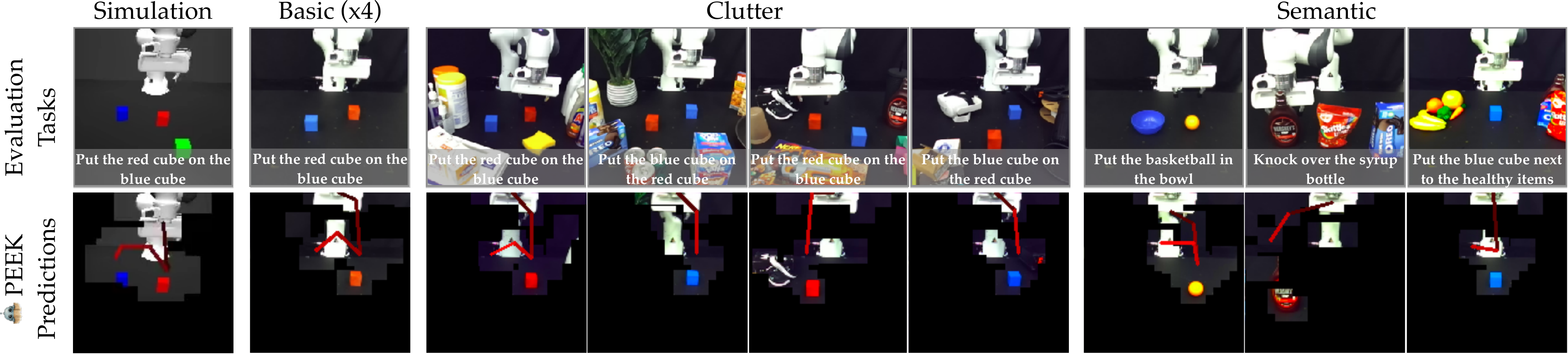}
    \caption{\footnotesize{\textbf{Franka Sim-to-Real Tasks.} Zero-shot evaluation environments along with associated path-drawn and masked images produced by \method. \task{Simulation} denotes the generated simulation data that the policies were trained on.}}
    \label{fig:exp:franka envs}
\end{figure*}

\begin{figure*}[tb]
    \centering
    \includegraphics[width=\linewidth]{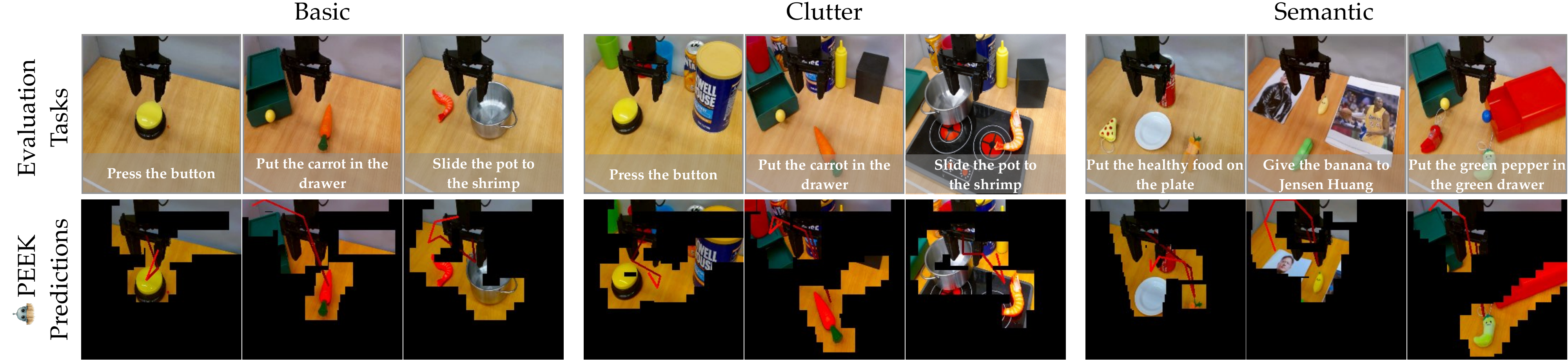}
    \caption{\footnotesize{\textbf{WidowX Tasks.} Evaluation environments along with associated path-drawn and masked images produced by \method.}}
    \label{fig:exp:bridge envs}
\end{figure*}

\section{Experimental Setup}
\label{sec:experiments}

To demonstrate the broad applicability of \method, we evaluate across two real-world robot embodiments, both 2D ($\pi_0$~\citep{black2024pi0}, \baseline{ACT}~\citep{zhao23aloha}) and 3D (\baseline{3DDA}~\citep{ke20243d}) policy classes, fine-tuning and training policies from scratch.
We evaluate zero-shot generalization from publicly available~\citep{walke2023bridgedata} and simulation-generated datasets to our custom setups, varying the task semantics and introducing visual clutter.

\noindent\textbf{Franka Sim-to-Real.} To study the semantic generalization and visual robustness induced by \method, we require a large-scale robotic dataset to cover all possible motions the policy might encounter during inference. Simulation offers a cheap, scalable approach to generate such a dataset without going through the effort of manual data collection.

We collect 2.5k trajectories of \emph{cube stacking} with a motion planner in MuJoCo environments with three colored cubes (sampled from \{red, green, blue, yellow\}) placed randomly on a $40\times40\text{cm}$ grid. See \Cref{fig:exp:franka envs} for a visualization of the data.
Our real-world setup consists of a Franka Emika Panda robot~\cite{grotz2024peract2,Zakka_Mink_Python_inverse_2025} with depth from processing RGB images from a Zed 2 stereo camera with FoundationStereo~\cite{wen2025foundationstereo}.

In the real world, we first test policy transfer on four fixed cube configurations (\task{Basic}), then add visual \task{Clutter} to assess visual robustness, and finally evaluate three \task{Semantic} tasks requiring reasoning about unseen objects and placements (\Cref{fig:exp:franka envs}). Each policy is evaluated for 5 trials per task, totaling 220 evaluations across 4 methods and 11 variations.

\noindent\textbf{WidowX BRIDGE.}
Our second environment uses a WidowX250 robot with a single Logitech C920 RGB camera, resembling the BRIDGE~\citep{walke2023bridgedata} environment, albeit without exactly reproducing camera angles and with a different table, objects, and background wall.
We re-label the BRIDGE-v2 dataset~\citep{walke2023bridgedata} (single camera angle) with \method\ according to \Cref{sec:method:policy} and zero-shot evaluate it on our setup.

We evaluate on a set of three tasks, representing basic generalization (to our custom robot setup), visualized in the \task{Basic} column of \Cref{fig:exp:bridge envs}. 
We then evaluate \task{Clutter}, which adds significant visual clutter to each of the three \task{Basic} tasks, and finally \task{Semantic}, representing difficult tasks that require visual-language reasoning to complete. We perform 5 evals per task with randomized object locations. 

\noindent\textbf{Baselines.}
In our Sim-to-Real experiments, we evaluate \method's application to 3D policies. We use \baseline{3DDA}~\citep{ke20243d} as our base policy and implement all baselines on top of it.
\begin{itemize}[itemsep=0pt, topsep=0pt, parsep=2pt,leftmargin=1em]
    \item \baseline{3DDA}~\citep{ke20243d}: A state-of-the-art language-conditioned 3D policy conditioned on depth, RGB, and language.
    \item \baseline{HAMSTER}~\citep{li2025hamster}: Fine-tunes a 13B parameter VLM to predict \emph{2D gripper paths} for a 3D policy to condition on. 
    \item \baseline{ARRO}~\citep{mirjalili2025augmentedrealityrobotsarro}: An \emph{explicit masking} baseline using GroundingDINO~\citep{liu2023grounding} to segment gripper and objects. 
\end{itemize}
We apply masks from \baseline{ARRO} and \baseline{\method} to both the RGB image and point clouds input to \baseline{3DDA}.

To show \method\ also applies to 2D policies of different architectures, we evaluate it on \baseline{ACT}~\citep{zhao23aloha} and \baseline{$\pi_0$}~\citep{black2024pi0}.
\begin{itemize}[itemsep=0pt, topsep=0pt, parsep=1pt,leftmargin=1em]
    \item \baseline{ACT}~\citep{zhao23aloha}: a small 90M parameter transformer policy we additionally condition with language embeddings~\citep{reimers-2019-sentence-bert}.
    \item \baseline{$\pi_0$}~\citep{black2024pi0}: A 3.5B parameter VLA first pre-trained on a large dataset, which we LoRA fine-tune on BRIDGE.
    \item \baseline{OTTER}~\citep{huang2025otter}: A 400M parameter transformer which \emph{implicitly masks} observations by discarding image patches with low CLIP-feature alignment to the task instruction.
    \item \baseline{ARRO}~\citep{mirjalili2025augmentedrealityrobotsarro}: \emph{Explicit masking} baseline introduced above.
\end{itemize}
We evaluate both \baseline{ARRO} and \baseline{\method} on top of both \baseline{ACT} and \baseline{$\pi_0$} as they are both policy-agnostic.

\section{Experimental Results}

Our evaluation aims to address the following questions: (Q1) How much does \method\ improve semantic and visual generalization across \textbf{diverse} policy architectures? (Q2) How accurately does \method\ help with \emph{where} and \emph{what}? and (Q3) How much does each component of \method\ contribute?
We answer these questions in order below.

\begin{figure*}[tb]
    \centering
    \includegraphics[width=\linewidth]{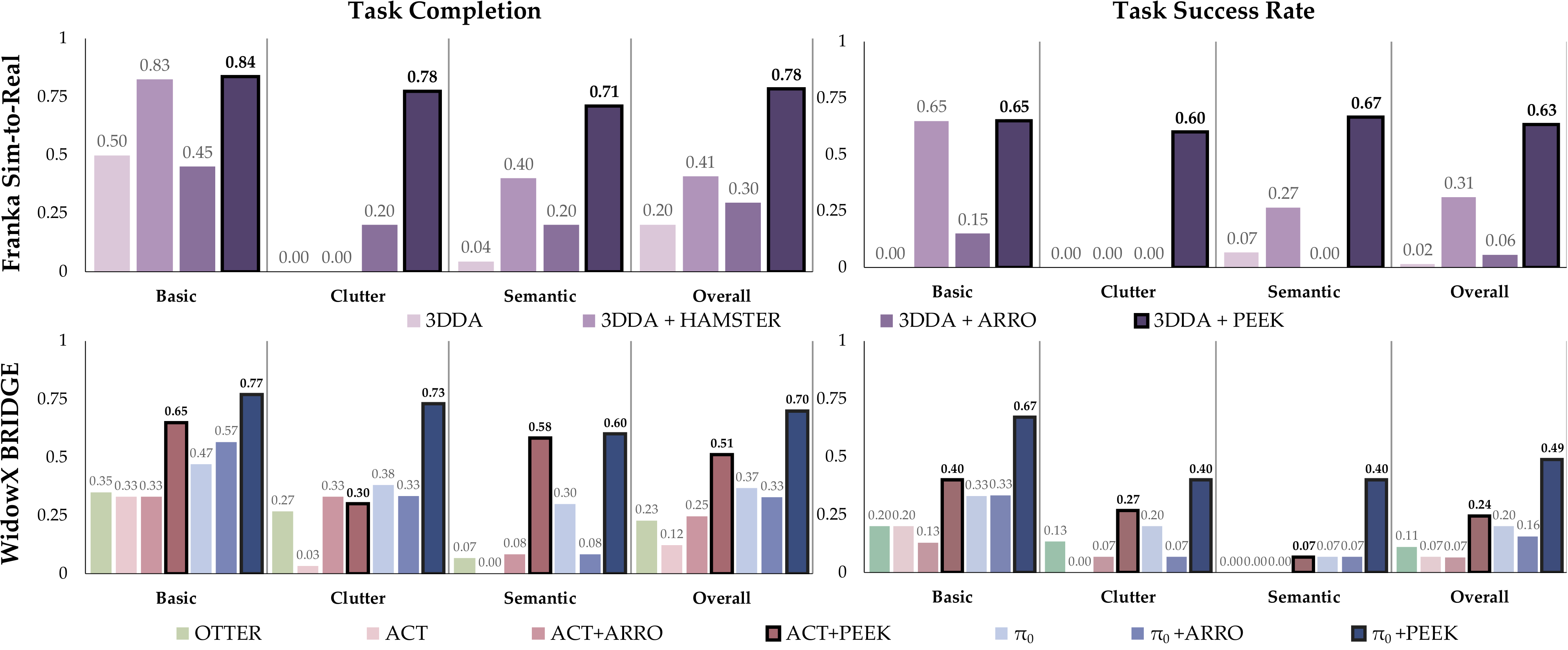}
    \caption{\footnotesize{\textbf{Real-World Zero-Shot Generalization Results.} Task completion rates (including partial credit for grasping or reaching objects correctly) and task success rates across 3 task variants: \task{Basic}, \task{Clutter}, and \task{Semantic} in our Franka Sim-to-Real experiments (top) and WidowX BRIDGE experiments (bottom). Results are averaged over all trials and tasks within each variant. \baseline{\method} results are bolded for visibility.} Full tables in Appendix \Cref{sec:apdx:tables}.}
    \label{fig:exp:real world results}
\end{figure*}

\subsection{Q1: Real-World Zero-Shot Generalization Experiments}
\noindent\textbf{Franka Sim-to-Real.}
We plot results in \Cref{fig:exp:real world results} (top). Overall, \baseline{3DDA+\method} improves vanilla \baseline{3DDA} by \textbf{41.4$\times$} and outperforms the best baseline, \baseline{3DDA+HAMSTER}, by \textbf{2$\times$} in overall success rates. 
While \baseline{HAMSTER} shows some semantic generalization via drawing paths ($40\%$ partial success on \task{Semantic}), it fails catastrophically when distractor objects are present in the scene ($0\%$ partial success on \task{Clutter}).
Instead, \baseline{\method}'s ability to also mask out irrelevant parts of the image usually completely hides task-irrelevant objects, allowing the policy to solve the task more often by only focusing on low-level control.

Meanwhile, \baseline{ARRO}, which masks-in the robot end-effector and task-relevant objects with pre-trained object detection models, often also includes task-irrelevant objects, confusing the \baseline{3DDA} policy.
\baseline{\method}'s VLM generalizes better to new objects and its paths help guide the policy even in cases where parts of irrelevant objects are included in the observation.
The baseline results demonstrate that answering only one of \emph{where} to focus or \emph{what} to do is not enough to achieve semantic generalization and visual robustness.
We visualize example \method\ VLM predictions in \Cref{fig:exp:franka envs}.

\noindent\textbf{WidowX BRIDGE.}
Next, we plot the WidowX results in \Cref{fig:exp:real world results} (bottom). 
\baseline{ACT+\method} and \baseline{$\pi_0$+\method} outperform their base models by \textbf{3.4$\times$} and \textbf{2.5$\times$} in overall success rates.
\baseline{ARRO} does not improve overall success rates of either base model as its pre-trained object detection module often fails to identify correct objects in clutter, and almost always fails to detect the robot gripper.
\baseline{\method}'s use of a VLM allows it to consistently mask-in the correct object and draw paths accurately starting from the gripper.
The VLM predictions visualized in \Cref{fig:exp:bridge envs} show how \baseline{\method}'s VLM provides effective paths and masks even in the face of distractors and tasks that require semantic and visual reasoning.

Meanwhile, \baseline{OTTER} performs poorly---better than \baseline{ACT} but worse than standard $\pi_0$, and far worse than either \baseline{\method} variation---\baseline{$\pi_0$+\method} overall achieves a \textbf{4.5$\times$} better success rate. 
This result highlights the importance of a policy-agnostic approach, such as \baseline{\method}, that can provide explicit path and mask guidance even to already strong base policies.

\subsection{Q2: Does \method\ answer the where and what?}
Comparing the first columns of Franka Sim-to-Real (\Cref{fig:exp:franka envs}) in \task{Simulation}, \task{Basic}, and \task{Clutter}, the benefits of paths and masking become apparent: %
masks remove distractors from the image---showing where to attend to---and the paths guide the policy to pick up the object---showing what to do.
Similar findings hold for the WidowX (\Cref{fig:exp:bridge envs}). 

While the masks tell the policy what to focus on, they alone are insufficient for solving semantic variation. Take, for example, the \task{Semantic} tasks; the policies' training data does not contain demonstrations featuring celebrities (``Give the banana to Jensen Huang'' in \Cref{fig:exp:bridge envs}) or various kinds of sweet treats (``Knock over the syrup bottle'', ``Put the blue cube next to the healthy items'' in \Cref{fig:exp:franka envs}). By letting the high-level VLM absorb the semantic generalization---proposing guiding paths---the policy can simply actualize the path into low-level actions to solve the task.

\subsection{Q3: How does each component contribute?}

\begin{wraptable}[8]{R}{0.5\columnwidth}
    \vspace{-0.3cm}
    \centering
    \resizebox{\linewidth}{!}{%
        \begin{tabular}{ccc}
        \toprule
        Paths $p$ & Masks $m$ & Success (\%) \\
        \midrule
        \xmark & \xmark & $33.5\pm3.1$ \\
        \cmark & \xmark & $52.8\pm2.9$ \\
        \xmark & \cmark & $65.6\pm3.1$ \\
        \cmark & \cmark & \textbf{$73.6\pm3.9$} \\
        \bottomrule
        \end{tabular}%
    }
    \caption{\footnotesize{Ablation of paths and masks on success rate.}}
    \label{tab:exp franka ablations}
\end{wraptable}

\noindent\textbf{Ablating Paths and Masks.}
We ablate the contributions of paths $p$ and masks $m$ on the performance of a language-conditioned 3D policy (\baseline{3DDA}) on the simulated cube stacking task in \Cref{tab:exp franka ablations}.
While the language-conditioned base policy can stack cubes, it often ignores instruction order, e.g., placing the blue cube on the red instead of the reverse. Adding only paths or only masks improves performance by $+19.3\%$ and $+32.1\%$, respectively. Masks outperform paths since they simplify the scene by removing the distractor cube, while paths alone leave ambiguity. Yet both remain limited: cube stacking highlights the insufficiency of purely predictive or minimal representations. Combining paths and masks, \method\ achieves gains of $+7.9\%$ over paths, $+20.8\%$ over masks, and $+40.1\%$ over the base policy.

\noindent\textbf{VLM Design Choices.}
To study VLM design choices, we evaluate on 1k holdout samples from BRIDGE-v2~\cite{walke2023bridgedata}, using Dynamic Time Warping (DTW) for paths~\cite{memmel2024strap} and Intersection over Union (IoU) for masks. 
Reducing the base model from 13B to 3B yields no loss in accuracy (both have DTW $0.12$, IoU $0.68$) while enabling faster closed-loop inference. Adding RoboPoint slightly improves these metrics and preserves semantic reasoning ability~\cite{li2025hamster}. Finally, joint prediction of paths and masks improves performance, giving a $+19.3\%$ relative gain over a mask-only model (IoU $0.57$) without degrading path accuracy (DTW $0.12$). Full results in Appendix \Cref{sec:apdx:vlm ablation details}.

Overall, we see that both paths and masks are essential to \method's ability to enhance \emph{policy} generalization, and our choice of a small VLM model that jointly predicts a unified path and mask representation great performance without sacrificing inference speed.

\section{Conclusion and Limitations}
We presented \textbf{\methodlong}, a framework that leverages VLMs to offload high-level reasoning in robot manipulation. By predicting point-based intermediate representations---paths that specify \emph{what} to do and masks that indicate \emph{where} to attend---\method\ provides policies with simplified, annotated observations, allowing them to focus on \emph{how} to act. Real-world evaluations demonstrate substantial improvements in zero-shot generalization across various policies. 

However, \method\ still inherits the biases and limitations of the underlying VLMs, which may fail in out-of-distribution scenarios or produce incorrect annotations. 
Our current representation is also limited to 2D point paths and masks; extending it to richer 3D or multimodal cues is an exciting direction. 
Moreover, although our annotation pipeline scales across existing robotics datasets, future work could explore how to bootstrap from a much broader corpus of video data.

\section*{Acknowledgements}
We thank Abrar Anwar for helping create the PEEK logo,
Helen Wang for lending us a difficult-to-obtain, official Labubu doll,
Raymond Yu for help setting up the initial BRIDGE table and FoundationStereo pipeline,
Markus Grotz for assistance in setting up the Franka controller stack (robits),
Yi Li for HAMSTER baseline help,
Andy Tang for assisting with initial BRIDGE camera alignment,
and William Chen for providing exact measurements for us to align the BRIDGE camera positions as best as possible.
We also thank Yondu.ai for hosting the Los Angeles Lerobot hackathon where we tried an early version of \method, and Yutai Zhou and Minjune Hwang for joining us in the competition.

Additionally, we acknowledge funding from the Army Research Lab and compute resources from the University of Southern California’s Center for Advanced Research Computing (CARC).
\printbibliography

\appendices

\begin{figure*}[h!]
    \centering
    \includegraphics[width=\linewidth]{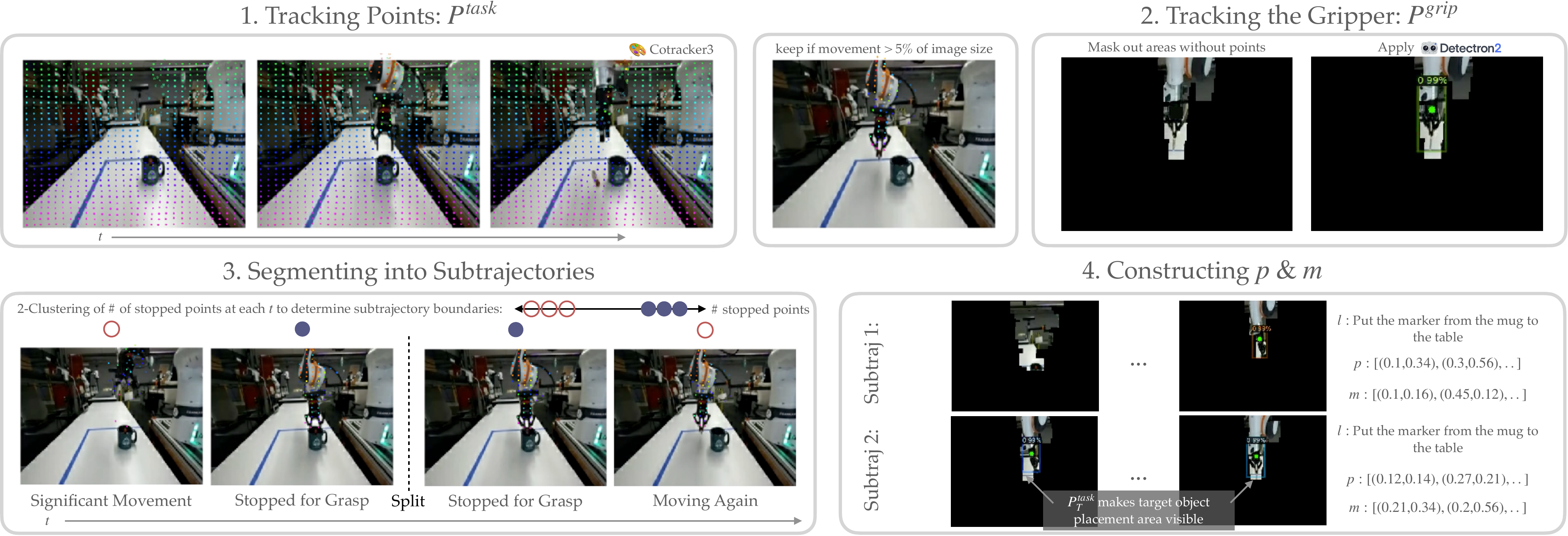}
    \caption{\footnotesize{\textbf{Data Labeling Pipeline.} A detailed overview of the data labeling pipeline as described in \Cref{apdx: data labeling}: We (1) use CoTracker3~\citep{karaev2025cotracker} to detect moving points across each trajectory, points are discarded if they do not move significantly, and the rest become task-relevant points $P^{task}$; (2) mask areas without $P^{task}$ to black and apply a pre-trained gripper detector to construct 2D gripper path points $P^{grip}$; (3) segment each trajectory into subtrajectories; and (4) construct gripper paths $p$ and task-relevant masking points $m$ for each subtrajectory. Notice in (4) that target object placement areas become visible beforehand through including points from the last timestep $P^{task}_T$.}}
    \label{fig:data labeling}
\end{figure*}
\section{}
\subsection{Data Annotation Pipeline Details}
\label{apdx: data labeling}

\noindent\textbf{Tracking Points.}
Given a trajectory of image observations $o_{1:T}$, we apply CoTracker3~\citep{karaev2025cotracker} to the video to track all moving points within the scene.
Points are initialized in a uniform grid across the entire pixel space (a $15\times15$ grid to $30\times30$ grid depending on how far objects are from the camera).
We initialize this point grid from the middle of the trajectory because in many datasets, the gripper is not visible at the first timestep.
Running CoTracker returns a set of all points and their normalized image locations across the trajectory.
We discard any point that does not move much throughout the trajectory, i.e., less than 5\% of the image size, because they are unlikely to be task-relevant.
The $N$ remaining points at each timestep $t$, $[\{(x, y)_i\}_{i=1}^N]_{t=1}^T$ with $(x, y) \in [0, 1]^2$, indicate both objects that move at some point during the trajectory and the robot gripper location.
Therefore, these are the \emph{task-relevant} points $P^{task}_{t}=\{(x,y)_{i=1}^N\}_t$.
See \Cref{fig:data labeling}~(1) for an overview of point tracking.

\noindent \textbf{Tracking the Gripper.}
To track the end-effector, we apply an object detection model (Detectron2~\citep{wu2019detectron2} fine-tuned by \citep{niu2024llarva} for end-effector detection) at every timestep. %

However, we found that na\"{i}vely applying the gripper detector resulted in noisy predictions that often did not include the robot.
To reduce the noise, we only keep pixels in $o_t$ around the significant points $P^{task}_{t}$, essentially masking out distractions irrelevant to the task.
We apply the detector to this reduced representation to obtain per-timestep gripper bounding boxes (filling in frames with no detected gripper with the average of adjacent detections) and average across them to obtain the end-effector points $P^{grip}_t=[(x_t,y_t)]_t$.
See \Cref{fig:data labeling}~(2) for a visual depiction.

\noindent\textbf{Segmenting into Subtrajectories.}
Because masks $m_t$ include points at the final timestep $T$, $m_t=P^{task}_{t}\cup P^{task}_{T}$, constructing $m_t$ from long-horizon trajectories can violate the minimality principle. For example, the policy does not have to know the placement location of an object until it has actually picked it up.
Therefore, we automatically break down the trajectories into \emph{subtrajectories}.
Our key insight is that when many task-relevant points \emph{stop moving}, the robot is likely to be manipulating an object. Vice versa, when those points start moving again, the robot is likely reaching or carrying an object.
This creates a natural approach to splitting a trajectory: by how many points in $P^{task}$ stop moving.

For each frame, $o_t$, we track how many points in $P^{task}_t$ don't move for the next 5 frames (3 for BRIDGE due to its higher control frequency). 
This creates a list of length $T$ containing the number of ``stopped points'' for each timestep, $t$.
On this list, we perform $K$-Means clustering with $K=2$, where the cluster with the smaller mean (fewer stopped points) corresponds to significant movement, e.g., the robot arm reaching an object, and the cluster with the larger mean corresponds to the robot performing fine-grained manipulation, e.g., grasping. 
Finally, we use these cluster assignments to find continuous sections $i, i+1, .., j - 1, j$ where the robot is manipulating an object, and use the middle frame of these sections $(j + i) / 2$ as subtrajectory split points.
This procedure results in a split of subtrajectories that end when the robot finishes a manipulation and start before it moves onto the next object manipulation. 
These subtrajectories create natural, shorter-horizon VLM prediction targets (see \Cref{fig:data labeling}~(3)).

\subsection{VLM Dataset Details}
\textbf{OXE.}
We augment the dataset by re-sampling the start and end of each subtrajectory within the first and last 20\% of the trajectory, repeating this procedure $5\times$. Additionally, we discard the first 20\% of each full trajectory as most of them don't feature the gripper or show very little movement.
Finally, we exclude FurnitureBench, Roboturk, Dobbe, BerkeleyCableRouting, LangTable, Kuka, and FMB.

\textbf{LIBERO-90.}
We re-render LIBERO-90~\citep{liu2023libero} to 256x256 from 128x128 following \citep{kim2024openvla}, consisting of 3958 successfully replayed and re-rendered demonstrations across 50 tasks.

\textbf{Postprocessing.}
Due to the autoregressive nature of transformers, the inference time grows linearly with the number of tokens predicted. To further reduce inference time, we follow \citep{li2025hamster} in reducing the number of points in $p_t$ and $m_t$ by applying the Ramer–Douglas–Peucker algorithm with tolerance thresholds $\epsilon=0.05$ and $\epsilon=0.1$, respectively.

\subsection{VLM and Policy Implementation Details}
\label{sec:apdx:implementation details}
For policy training, path and masks are labeled with VLM queries every $H=30$ and $H=32$ timesteps for BRIDGE and Franka Sim-to-Real.
During rollouts, \method's VLM is queried every $H=25$ and $H=32$ timesteps for BRIDGE and Sim-to-Real respectively, 
and the policies all predict action chunks of length 5 and 8 respectively.

\subsection{VLM Ablation Results}
\label{sec:apdx:vlm ablation details}
\begin{table*}[htbp]
\centering
\resizebox{\textwidth}{!}{%
    \centering
    \begin{tabular}{c c c c c c c}
    \toprule
    Model Size & Prediction Target & Training Dataset(s) & DTW Distance $\downarrow$ ($p$) & First Point L2 $\downarrow$ ($p$) & Last Point L2 $\downarrow$ ($p$) & IoU $\uparrow$ ($m$) \\
    \midrule
    3B  & $p$      & OXE+RoboPoint              & 0.1239 & 0.0445 & 0.1479 & N/A     \\
    13B & $p$      & OXE+RoboPoint              & 0.1197 & 0.0450 & 0.1447 & N/A     \\
    3B  & $m$      & OXE+RoboPoint              & N/A     & N/A     & N/A     & 0.5799 \\
    13B & $m$      & OXE+RoboPoint              & N/A     & N/A     & N/A     & 0.5858 \\
    3B  & $p$+$m$ & OXE+RoboPoint             & 0.1229 & 0.0426 & 0.1422 & 0.6863 \\
    13B & $p$+$m$ & OXE+RoboPoint              & 0.1219 & 0.0419 & 0.1491 & 0.6885 \\
    \midrule
    3B  & $p$+$m$ & OXE (inc. BRIDGE)        & 0.1237 & 0.0426 & 0.1467 & 0.6785 \\
    13B & $p$+$m$ & OXE (inc. BRIDGE)        & 0.1222 & 0.0435 & 0.1494 & 0.6835 \\
    3B  & $p$+$m$ & BRIDGE+RoboPoint           & 0.1249 & 0.0444 & 0.1430 & 0.6792 \\
    13B & $p$+$m$ & BRIDGE+RoboPoint           & 0.1208 & 0.0428 & 0.1459 & 0.6855 \\
    3B  & $p$+$m$ & BRIDGE & 0.1237 & 0.0435 & 0.1438 & 0.6727 \\
    13B & $p$+$m$ & BRIDGE & 0.1263 & 0.0437 & 0.1447 & 0.6798 \\
    \bottomrule
    \end{tabular}
    }
    \caption{\textbf{VLM Ablations.} Evaluation on 1000 hold-out BRIDGE dataset samples for paths and masks from our data labeling pipeline compared across VLM model size (3B and 13B parameters), prediction target (path $p$, mask $m$), and training datasets (OXE, BRIDGE, RoboPoint). The top half of the table ablates the model and prediction target, the bottom half ablates the training dataset.}
    \label{tab:vlm_ablations}
\end{table*}

We ablate the VLM model size (3B and 13B parameters), training dataset mixtures (OXE labeled with \Cref{apdx: data labeling} (OXE), the BRIDGE training split from our labeled OXE (BRIDGE), and the RoboPoint dataset~\cite{yuan2024robopoint}), and prediction target ($p$, $m$, $p+m$) in \Cref{tab:vlm_ablations}. The metrics recorded are DTW distance (DTW L2 distance between predicted and ground truth $p$), First Point L2 (L2 distance between the first point in predicted and ground truth $p$), Last Point L2 (L2 distance between the last point in predicted and ground truth $p$) for path $p$ predictions, intersection over union (IoU) for mask predictions $m$. Models are evaluated on 1k holdout samples from the BRIDGE test split from our labeled OXE.

Overall, there is a minimal difference in performance between the 3B and 13B parameter models; hence, we chose to use the 3B parameter VILA model for \method\ for its faster inference speed. 
The combination of predicting both paths and masks with the same model improves the performance on paths alone or masks alone on the 3B parameter model. 
Finally, including the full OXE dataset and including RoboPoint VQA/Pointing data overall helps performance on the BRIDGE evaluation dataset over just using BRIDGE alone.

\subsection{Full Results Tables}
\label{sec:apdx:tables}
We display full results tables for the Franka Sim-to-Real experiments in \Cref{tab:appendix:full sim2real results} and for the BRIDGE experiments in \Cref{tab:appendix:full bridge results}.
\begin{table*}[htbp]
\centering
\begin{subtable}{\textwidth}
\resizebox{\textwidth}{!}{%
\begin{tabular}{l c c c c}
\toprule
\textbf{Basic Tasks} & 3DDA & 3DDA+HAMSTER & 3DDA+ARRO & 3DDA+\method \\
\midrule
Put the red cube on the blue cube & 0.50 & 0.95 & 0.25 & 1.00 \\
Put the blue cube on the red cube & 0.50 & 1.00 & 0.20 & 0.85 \\
Put the red cube on the blue cube & 0.50 & 0.40 & 0.80 & 0.70 \\
Put the blue cube on the red cube & 0.50 & 0.80 & 0.35 & 0.80 \\
\textbf{Average} & 0.50 & 0.82 & 0.45 & 0.83 \\
\midrule
\textbf{Vis \& Obj Clutter} & & & & \\
\midrule
Put the red cube on the blue cube & 0.00 & 0.00 & 0.10 & 1.00 \\
Put the blue cube on the red cube & 0.00 & 0.00 & 0.00 & 0.70 \\
Put the red cube on the blue cube & 0.00 & 0.00 & 0.50 & 0.60 \\
Put the blue cube on the red cube & 0.00 & 0.00 & 0.20 & 0.80 \\
\textbf{Average} & 0.00 & 0.00 & 0.20 & 0.77 \\
\midrule
\textbf{Semantic} & & & & \\
\midrule
Knock over the syrup bottle   & 0.20 & 0.20 & 0.50 & 0.80 \\
Put the basketball in the bowl   & 0.00 & 0.7 & 0.00 & 0.60 \\
Put the blue cube next to the healthy items & 0.00 & 0.13 & 0.26 & 0.80 \\
\textbf{Average} & 0.04 & 0.40 & 0.20 & 0.71 \\
\bottomrule
\end{tabular}
}
\caption{Partial completion rates per task.}
\end{subtable}

\hfill

\begin{subtable}{\textwidth}
\resizebox{\textwidth}{!}{%
\begin{tabular}{l c c c c}
\toprule
\textbf{Basic Tasks} & 3DDA & 3DDA+HAMSTER & 3DDA+ARRO & 3DDA+\method \\
\midrule
Put the red cube on the blue cube & 0.00 & 0.80 & 0.20 & 1.00 \\
Put the blue cube on the red cube & 0.00 & 1.00 & 0.20 & 0.60 \\
Put the red cube on the blue cube & 0.00 & 0.40 & 0.20 & 0.40 \\
Put the blue cube on the red cube & 0.00 & 0.40 & 0.20 & 0.60 \\
\textbf{Average} & 0.00 & 0.65 & 0.15 & 0.65 \\
\midrule
\textbf{Vis \& Obj Clutter} & & & & \\
\midrule
Put the red cube on the blue cube & 0.00 & 0.00 & 0.10 & 1.00 \\
Put the blue cube on the red cube & 0.00 & 0.00 & 0.00 & 0.70 \\
Put the red cube on the blue cube & 0.00 & 0.00 & 0.50 & 0.60 \\
Put the blue cube on the red cube & 0.00 & 0.00 & 0.20 & 0.80 \\
\textbf{Average} & 0.00 & 0.00 & 0.20 & 0.77 \\
\midrule
\textbf{Semantic} & & & & \\
\midrule
Knock over the syrup bottle   & 0.20 & 0.20 & 0.00 & 0.60 \\
Put the basketball in the bowl   & 0.00 & 0.60 & 0.00 & 0.60 \\
Put the blue cube next to the healthy items & 0.00 & 0.00 & 0.00 & 0.80 \\
\textbf{Average} & 0.06 & 0.26 & 0.00 & 0.71 \\
\bottomrule
\end{tabular}
}
\caption{Success rates per task.}
\end{subtable}
\caption{\textbf{Franka Sim-to-Real Results Table.} All task success/completion rates for each baseline are averaged over 5 trials.}
\label{tab:appendix:full sim2real results}
\end{table*}
\begin{table*}[htbp]
\centering
\begin{subtable}{\textwidth}
\resizebox{\textwidth}{!}{%
\begin{tabular}{l c c c c c c c}
\toprule
\textbf{Basic Tasks} & OTTER & ACT & ACT+ARRO & ACT+PEEK & $\pi_0$ & $\pi_0$+ARRO & $\pi_0$+PEEK \\
\midrule
Slide the pot to shrimp & 0.20 & 0.70 & 0.40 & 0.70 & 0.50 & 0.50 & 1.00 \\
Push the button         & 0.70 & 0.30 & 0.40 & 0.80 & 0.50 & 0.60 & 0.50 \\
Put carrot in drawer    & 0.15 & 0.00 & 0.20 & 0.45 & 0.40 & 0.60 & 0.80 \\
\textbf{Average}        & 0.35 & 0.33 & 0.33 & 0.65 & 0.33 & 0.57 & 0.77 \\
\midrule
\textbf{Obj \& Vis Clutter} & & & & & & & \\
\midrule
Slide the pot to shrimp & 0.20 & 0.10 & 0.30 & 0.50 & 0.40 & 0.50 & 0.70 \\
Push the button         & 0.60 & 0.00 & 0.40 & 0.70 & 0.40 & 0.10 & 0.80 \\
Put carrot in drawer    & 0.00 & 0.00 & 0.30 & 0.55 & 0.20 & 0.65 & 0.70 \\
\textbf{Average}        & 0.27 & 0.03 & 0.33 & 0.58 & 0.33 & 0.42 & 0.73 \\
\midrule
\textbf{Semantic} & & & & & & & \\
\midrule
Put pepper in box       & 0.00 & 0.00 & 0.10 & 0.65 & 0.50 & 0.05 & 0.60 \\
Give banana to Jensen   & 0.10 & 0.00 & 0.15 & 0.15 & 0.20 & 0.20 & 0.70 \\
Put food on plate       & 0.10 & 0.00 & 0.00 & 0.10 & 0.20 & 0.00 & 0.50 \\
\textbf{Average}        & 0.07 & 0.00 & 0.08 & 0.30 & 0.30 & 0.08 & 0.60 \\
\bottomrule
\end{tabular}

}
\caption{Partial completion rates per task.}
\end{subtable}

\hfill %

\begin{subtable}{\textwidth}
\resizebox{\textwidth}{!}{%
\begin{tabular}{l c c c c c c c}
\toprule
\textbf{Basic Tasks} & OTTER & ACT & ACT+ARRO & ACT+PEEK & $\pi_0$ & $\pi_0$+ARRO & $\pi_0$+PEEK \\
\midrule
Slide the pot to shrimp & 0.00 & 0.40 & 0.00 & 0.40 & 0.20 & 0.40 & 1.00 \\
Push the button         & 0.60 & 0.20 & 0.40 & 0.60 & 0.40 & 0.40 & 0.40 \\
Put carrot in drawer    & 0.00 & 0.00 & 0.00 & 0.20 & 0.40 & 0.20 & 0.60 \\
\textbf{Average}        & 0.20 & 0.20 & 0.13 & 0.40 & 0.33 & 0.33 & 0.67 \\
\midrule
\textbf{Obj \& Vis Clutter} & & & & & & & \\
\midrule
Slide the pot to shrimp & 0.00 & 0.00 & 0.00 & 0.40 & 0.20 & 0.20 & 0.40 \\
Push the button         & 0.40 & 0.00 & 0.20 & 0.40 & 0.20 & 0.00 & 0.60 \\
Put carrot in drawer    & 0.00 & 0.00 & 0.00 & 0.00 & 0.20 & 0.40 & 0.20 \\
\textbf{Average}        & 0.13 & 0.00 & 0.07 & 0.27 & 0.20 & 0.20 & 0.40 \\
\midrule
\textbf{Semantic} & & & & & & & \\
\midrule
Put pepper in box       & 0.00 & 0.00 & 0.00 & 0.20 & 0.00 & 0.00 & 0.20 \\
Give banana to Jensen   & 0.00 & 0.00 & 0.00 & 0.00 & 0.00 & 0.20 & 0.60 \\
Put food on plate       & 0.00 & 0.00 & 0.00 & 0.00 & 0.20 & 0.00 & 0.40 \\
\textbf{Average}        & 0.00 & 0.00 & 0.00 & 0.07 & 0.07 & 0.07 & 0.40 \\
\bottomrule
\end{tabular}

}
\caption{Success rates per task.}
\end{subtable}
\caption{\textbf{WidowX BRIDGE Results Table.} All task success/completion rates for each baseline are averaged over 5 trials.}
\label{tab:appendix:full bridge results}
\end{table*}

\end{document}